# SELF-ADAPTATION MECHANISM TO CONTROL THE DIVERSITY OF THE POPULATION IN GENETIC ALGORITHM

Chaiwat Jassadapakorn[1] and Prabhas Chongstitvatana[2]

Department of Computer Engineering, Chulalongkorn University, Bangkok, Thailand
[1]chaiwat.ja@student.chula.ac.th, [2]prabhas@chula.ac.th

## ABSTRACT

*One of the problems in applying Genetic Algorithm is that there is some situation where the evolutionary process converges too fast to a solution which causes it to be trapped in local optima. To overcome this problem, a proper diversity in the candidate solutions must be determined. Most existing diversity-maintenance mechanisms require a problem specific knowledge to setup parameters properly. This work proposes a method to control diversity of the population without explicit parameter setting. A self-adaptation mechanism is proposed based on the competition of preference characteristic in mating. It can adapt the population toward proper diversity for the problems. The experiments are carried out to measure the effectiveness of the proposed method based on nine well-known test problems. The performance of the adaptive method is comparable to traditional Genetic Algorithm with the best parameter setting.*

## 1. INTRODUCTION

Genetic Algorithm (GA) is a probabilistic search and optimization algorithm. The GA begins with a random population -- a set of solutions. A solution (or an individual) is represented by a fixed-length binary string. A solution is assigned a fitness value that indicates the quality of solution. The high-quality solutions are more likely to be selected to perform solution recombination. The crossover operator takes two solutions. Each solution is split in two pieces. Then, the four pieces of solutions are exchanged to reproduce two solutions. The population size is made constant by substituting some low-quality solutions with these new solutions. The selection process is an important factor of the success of GA. The application of GA is numerous, for example, it is used to optimize the parameters and topology of the network [1].

An important issue in applying GA to solve problems is a phenomenon called premature convergence [2]. It is the situation that an evolutionary process is converged too fast to a solution (or a few solutions) which causes it to be trapped in a local optima. The most common cause of premature convergence is the lack of diversity coupled with ineffectiveness of the crossover operator to search for a new solution. Without an adequate diversity a few better individuals dominate the population in a short period of time. When the population diversity is lost, the evolutionary process cannot progress. This is because some necessary genetic materials, which may be the part of solution, are lost.

To improve the performance of GA, many works proposed enhanced strategies by embedding the diversity-maintenance feature in different forms. Most of these works are reviewed in the next section. Unfortunately, they require a priori knowledge to setup parameters that affect the

                                    

degree of population diversity in the evolutionary process. Setting parameters incorrectly leads to unsuitable population diversity for the problem and causes poor performance.

In this work, a method to control diversity of a population without explicit parameter setting is proposed. The main idea is to regard the population as a multi-racial society where a group of similar chromosomes represents a race. When recombination occurs within a race the diversity of the population will be low. Vice versa when recombination occurs between different races the diversity will be high. To control the diversity, the selection criteria for mating include the *difference function* (defined in section 3) which measures dissimilarity of two individuals in addition to traditional fitness values. This function determines (with a fine degree) a selection of similar or dissimilar mate, according to our analogy selecting a dissimilar mate acts like a marriage across races. This will affect the diversity of the population. The self-adaptation mechanism comes from the observation that by letting the more successful marriage (defined by the success of their offspring) prospered the wellness of the society will increase. A proposed measurement called *contribution* (defined in section 4) is used to monitor the effect of recombination. With this indication, a decision can be made to adapt the type of recombination as needed. The experiments are carried out to measure the effectiveness of the proposed method using nine well-known test problems in the literature.

The organization of this paper is as follows. The next section reviews the related work. After that, the mating method and the proposed diversity control are explained. The use of the test problems is described. The details of our experiment are presented, and finally, the conclusion is given.

## 2. RELATED WORK

Population diversity is still an active research in Evolutionary Computation. To measure the diversity of population, many distance functions are proposed. The two individuals are close in distance if they are similar. Conversely, the large distance means the two individuals are different. Since an individual chromosome in GA is represented with a binary string, the Hamming distance is widely used.

To maintain the diversity of population, many strategies are proposed. The sharing method [3] is the most frequently used technique for maintaining population diversity. It is inspired by natural ecosystem. Each individual is forced to share its fitness value to its neighbors. The survival probability of an individual depends on its fitness value and its difference from others in the neighborhood. This approach encourages the exploration of the new region in a solution space.

The ranked space method [4] is another strategy. This approach embeds the diversity-maintaining mechanism explicitly by the use of two ranks in the selection process called the quality rank and the diversity rank. The combination of these two ranks is used to influence the selection probability. With this approach, the fitter individual is selected and at the same time the population diversity is maintained.

Another strategy is the approach called restricted mating. The restricted mating applies conditions such as restriction or encouragement, to select an individual and its mate partner. For example, the difference between pairs measured by the Hamming distance is used [5].

In [6], a selection scheme inspired by the animal mating behavior is proposed. This method applies dissimilar measurement to the pair of individuals. The first mate is selected with a traditional scheme--the higher fitness value, the more chance to be selected. The second mate is selected by considering another feature which can be dependent on the first partner (this process is called seduction). Subsequently, in [7], the chance to be selected as the second partner is affected from the combination of the fitness value and the difference from the first partner. Since the mating procedure depends on the difference in each pair of individuals, it can maintain the population diversity in an indirect way.

Other works that study the used of restricted mating are [8, 9]. In the work [8], the tabu multi-parent genetic algorithm (TMPGA) is presented. The mating of multiple parents in TMPGA is restricted by the strategy of tabu search. The tabu list is used for preventing incest and maintaining the diversity of population. In the work [9], a restricted mating is used for real-coded GA incorporating with other operators.

The DCGA (diversity control oriented genetic algorithm) [10] is another well-known strategy. The population of the next generation in DCGA is the merging of a previous generation population and their offspring with duplicate removal and sorting. The CPSS (cross-generational probabilistic survival selection) operator is applied to each individual. This causes the best individual to be selected. The next individuals are selected by chance. The higher difference from the best individual gives that individual higher probability to be selected. If the number of individual is insufficient after apply the CPSS to the whole population, the new random individuals are generated.

The next strategy is CSGA (complementary surrogate genetic algorithm) [11]. It is widely known that applying the mutation operator can preserve the diversity. However, CSGA has the diversity-maintenance feature without an explicit mutation. The distinguished feature of the CSGA is the inclusion of complementary surrogate set (CSS) into the population. The CSS is an individual or a set of individuals adding to the population for guaranteeing that each bit position of the whole population is diverse (not all '0' or all '1').

Another strategy is the selection scheme called FUSS (fitness uniform selection scheme) [12]. Let the lowest and highest fitness values in the population be $f_{min}$ and $f_{max}$ respectively. The FUSS will select a fitness $f$ uniformly in the interval $[f_{min}, f_{max}]$. Then the individual with fitness value nearest to $f$ is selected. The FUSS maintains a diversity better than a standard selection scheme since a distribution over the fitness value is used. Therefore, the higher and the lower fitness individuals are mixed in the selection. The other forms of selection scheme are proposed in [13, 14].

The multiploid genetic algorithm [15] is another mechanism. It was found that many life forms in nature are compound with more than one chromosome (multiploid) together with some mechanism for determining the gene expression. This observation can be used to implement an alternate GA. The multiploid GA provides a diversity-maintenance feature. This useful diversity preservation is suitable for many problems. However, the better result comes at the expense of extra computational time and space usage. The other techniques to maintain population diversity are duplicate genotype removal [16], mutation operator [17], adaptive crossover and mutation [18, 19], and parallel system [20], for example.

The population diversity maintenance is widely researched in multi-objective evolutionary algorithm (MOEA) area [17, 20-23]. In the works [22, 23], genetic diversity is explicitly preserving by considering it as an addition objective in the evaluation phase. The results show that this technique is effective at converging towards the Pareto-optimal set and distributing the population along it.

## 3. MATING

The main idea of the proposed diversity control is the competition between groups of individual with different degree of diversity. Therefore, it requires a mechanism to differentiate characteristic of diversity in each group. A parameter which affect characteristic of diversity in a mating is called *preference type*. The value of preference type indicates the preference of individual to recombine with the individual that is different from it. The higher value of preference type leads to the creation of offspring which differ from the parent. Conversely, the lower value of preference type leads to the creation of offspring which are close to the parent. Therefore, the offspring are not encouraged to be different. The detail of selection process in the mating is given below.

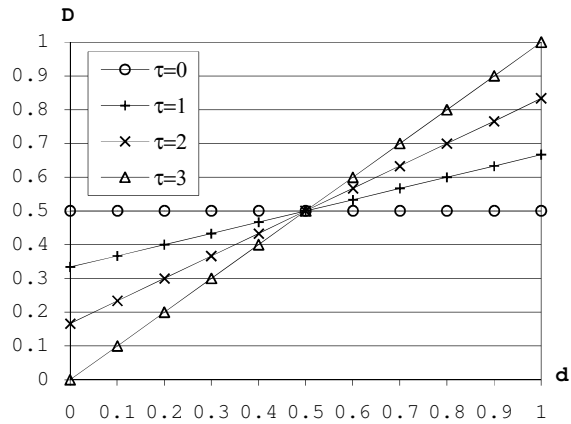

Figure 1. An example of a difference function (equation 2).

For maximization problems with non-negative fitness value for all possible solutions, given the first selected individual $x_1$ which is selected by a traditional selection method (tournament selection is used in the experiment), the preference type assigned to $x_1$ is used to calculate the chance of another individual to be selected as its partner. Let $d$ represents the difference between the first selected individual and a candidate, $\tau$ represents the preference type, and $D$ represents a function of $d$ and $\tau$, called the *difference function*. The candidate who has a higher $D$ value has more chance to be selected as the second partner. The selection criterion depends on the difference function and the fitness value (equation 1).

$$x_2 = \underset{i \in s_t}{\operatorname{argmax}}[f(y_i) \cdot D(\tau, d_i)] \tag{1}$$

$x_2$ represents the selected partner, $y_i$ is the $i^{\text{th}}$ candidate which are randomly selected from the population, $f$ is the fitness function, and $s_t$ is the tournament size. In the experiment, a basic difference function is used (equation 2).

$$D(\tau, d_i) = 0.5 + \frac{\tau}{\tau_{max}}(d_i - 0.5) \tag{2}$$

$\tau_{max}$ is the maximum preference type, $0 \leq \tau \leq \tau_{max}$, and $d_i$ is the difference between the first selected individual and the candidate $y_i$:

$$d_i = \frac{h(y_i, x_l)}{l} \tag{3}$$

$h$ is the Hamming distance of two individuals, and $l$ is the length of chromosome. An example of relationship between $D$, $d$ and is shown in Figure 1.

Please note that when $\tau$ is 0, the probability of selection does not depend on $d$. This is equivalent to a traditional selection method where the chance to be selected depends only on the fitness value. The higher value of $\tau$ gives more weight to the difference between individuals, which influences the population towards more diversity.

The concept of the proposed mating is related to the traditional restricted mating method but instead of using a heuristic to qualify a mate, it characterises the qualification of a mate using the difference function.

## 4. DIVERSITY CONTROL

The proposed mating is applied to construct a diversity control procedure that has capability to adapt the degree of diversity in a population to suite a given problem. Since the degree of diversity is controlled by the specification of preference type, the goal of this diversity control procedure is to search for a suitable preference type for a given problem. The idea is to use multiple preference types for the mating process. Each preference type is used equally first. The effectiveness of each preference type is evaluated, that causes the change of the frequency of use in the next iteration. This leads to the design of measurement called *contribution*.

Contribution is a measurement of the merit of each preference type in the term of how often they construct better individuals in the next generation population.

$$Contribution(\tau, t) = \frac{\# SuccCross(\tau, t)}{\# Cross(\tau, t)} \tag{4}$$

As shown in equation 4, the contribution of each preference type $\tau$ at the generation $t$ can be calculated by two terms: the number of successful crossover (denoted by $\# SuccCross(\tau, t)$) and the number of crossover times (denoted by $\# Cross(\tau, t)$). The successful crossover is the one that produces at least one child who is better than both parents considered by the fitness value. This term is normalized by the total number of crossover of each preference type.

The contribution is the measurement of the ratio of the better individual creation. The preference type with higher contribution indicates the higher effectiveness.

With the use of contribution measurement, the preference types are in direct competition against each other to be used. The preference type that performs well will be promoted. The preference type that is inferior will be demoted. The promoted (demoted) mechanism causes increasing (decreasing) the chance for a preference type to be used in proportion to its contribution. This scheme leads to the concentration of computational effort to the promising preference type which causes the adaptation of diversity for a given problem.

The process of the diversity control system can be summarized as follows:

1.  Randomly generate the population of individual.
2.  Evaluate each individual by a fitness function.

3.  Set the contribution equally for each preference type for the first time.
4.  Select an individual and its partner with the proposed mating procedure. The probability of choosing a preference type is proportional to its contribution.
5.  Reproduce two new individuals for the next generation by crossover.
6.  Repeat step 4 and 5 for the whole population.
7.  Evaluate each new individual by the fitness function.
8.  Compare the fitness value of the new individuals and their parental individuals. Calculate contribution of each preference type.

Repeat step 4-8 until reach the final generation.

Table 1. The test problems.

| Function name | Function | Parameters |
|---|---|---|
| One-max | $f_1(x) = \sum_{i=1}^{n} x_i$ | $x_i \in \{0,1\}$, $n = 45$, $l = 45$ |
| Deceptive function | $f_2(x) = \sum_{i=1}^{n} g(x_i)$ $g(x_i) = \begin{cases} 0.9 & if & |x_i| = 0 \\ 0.8 & if & |x_i| = 1 \\ 0 & if & |x_i| = 2 \\ 1 & if & |x_i| = 3 \end{cases}$ | $x_i$ is a 3-bit string, $|x_i|$ is the number of bit '1' in a 3-bit string, $n = 15$, $l = 45$ |
| Multimodal function [24] | $f_3(x) = e^{-2(\ln 2)\left(\frac{x-0.08}{0.854}\right)^2} \sin^6[5\pi(x^{3/4} - 0.05)]$ | $x_i \in [0,1]$, $l = 30$ |
| De Jong's f1 (Sphere) | $f_4(x) = \sum_{i=1}^{n} x_i^2$ | $x_i \in [-5.12, 5.11]$, $n = 3$, $l = 30$ |
| De Jong's f3 (Step) | $f_5(x) = \sum_{i=1}^{n} \lfloor x_i \rfloor$ | $x_i \in [-5.12, 5.11]$, $n = 5$, $l = 50$ |
| Shaffer's f6 | $f_6(x) = 0.5 + \frac{(\sin\sqrt{x_1^2 + x_2^2})^2 - 0.5}{(1 + 0.001(x_1^2 + x_2^2))^2}$ | $x_i \in [-102.4, 102.3]$, $l = 22$ |
| Rastrigin | $f_7(x) = 10 \cdot n + \sum_{i=1}^{n} (x_i^2 - 10 \cdot \cos(2 \cdot \pi \cdot x_i))$ | $x_i \in [-5.12, 5.11]$, $n = 3$, $l = 30$ |
| Schwefel | $f_8(x) = \sum_{i=1}^{n} (-x_i \cdot \sin(\sqrt{|x_i|}))$ | $x_i \in [-512, 511]$, $n = 3$, $l = 30$ |
| Griewangk | $f_9(x) = \sum_{i=1}^{n} \frac{x_i^2}{4000} - \prod_{i=1}^{n} \cos\left(\frac{x_i}{\sqrt{i}}\right) + 1$ | $x_i \in [-512, 511]$, $n = 3$, $l = 30$ |

## 5. EXPERIMENT

The proposed method is evaluated using the well-known test problems in GA. They require different degree of diversity in the population to solve them efficiently. The test problems are summarized in Table 1 (where $l$ is the length of chromosome of each problem). The plots of function $f_3$ to $f_9$ are shown in the appendix. Since $f_4$ to $f_9$ are minimization problems, they can be processed as maximization problems by simply convert to new functions as equation 5.

$$f^*(x) = f_{max} - f(x) \qquad (5)$$

$f^*(x)$ is the result function for maximization, $f_{max}$ is the maximum function value in the given range, and $f(x)$ is the original function. The value of the new function $f^*(x)$ is non-negative for all $x$ within the given range.

The performance of the proposed method is compared with the non-adaptive one and the DCGA since it is one of the most well-known methods to maintain population diversity in GA. The non-adaptive procedure is GA using varying of mutation rate ($P_m$) which is the diversity maintaining method using in traditional GA. The parameters used in the experiment are shown in Table 2. The one-point crossover is used in the experiment. The mutation operation in the proposed method is excluded. This emphasises that the diversity in the population comes from the use of mating by preference types only.

Table 2.  The parameters used in the experiment.

| Parameter | Value |
|---|---|
| Population size | 400 |
| Length of chromosome | 22-50 bits |
| Number of generation | 200 |
| Number of repeated run | 500 |
| Crossover probability ($P_c$) | 100% |
| Mutation rate ($P_m$) (non-adaptive only) | 0.00-0.05 (0-5%) |
| Tournament size | 3 |
| Number of preference type (proposed method only) | 4 ($\tau$ = 0-3) |

## 6. COMPARE TO NON-ADAPTIVE PROCEDURE

To compare the performance for solving problems between adaptive and non-adaptive procedure, the computational effort [25] is used as the indicator. The computational effort is defined as the average number of individual to be evaluated to obtain the solution.

Let $P(M,i)$ be the probability of finding the solution within the generation $i$, $M$ is the number of individual in the population. $P$ can be observed by repeating the experiment many times. $R(M,i,z)$ be the number of run required to find the solution in the generation $i$ with the confidence $z$. $R(M,i,z) = \lceil log(1-z) / log(1- P(M,i)) \rceil$. The minimum number of individual that must be processed to find the solution with the confidence $z$ is $I(M,i,z) = M \times i \times R(M,i,z)$. The minimum value of $I(M,i,z)$ is defined as the computation effort. The confidence $z$ in this work is 99%. $I^*$ is the generation that the minimum effort occurs. $N$ is the number of run that found the solution.

Table 3 shows the computational efforts of the non-adaptive and the adaptive procedures for the test problems. The table shows the good performance of the proposed method comparing to the non-adaptive one. The computational effort scores are comparable with the best non-adaptive procedure (GA with the best setting of mutation rate). For deceptive function, De Jong's f1 function, Shaffer's f6 function, and Rastrigin function, the proposed method is superior to the non-adaptive one, since the computational effort scores are lower than the best non-adaptive procedure.

Table 3. Comparing the computational efforts of proposed method and the non-adaptive procedures (denoted as $P_m$=0.00 to 0.05).

| Problem | Method | $\overline{\text{Gen}}$ | $I^*$ | Effort | N |
|---|---|---|---|---|---|
| One-Max | $P_m$=0.00 | 11.44 | 14 | 6,000 | 500 |
| | $P_m$=0.01 | 11.89 | 15 | 6,400 | 500 |
| | $P_m$=0.03 | 15.17 | 19 | 8,000 | 500 |
| | $P_m$=0.05 | 29.35 | 64 | 26,000 | 500 |
| | proposed | 13.01 | 16 | 6,800 | 500 |
| Deceptive function | $P_m$=0.00 | 18.69 | 22 | 64,400 | 265 |
| | $P_m$=0.01 | 30.75 | 32 | 92,400 | 278 |
| | $P_m$=0.03 | 150.01 | 191 | 307,200 | 359 |
| | $P_m$=0.05 | - | - | - | 0 |
| | proposed | 23.96 | 30 | 24,800 | 472 |
| Multimodal function | $P_m$=0.00 | 20.26 | 26 | 172,800 | 140 |
| | $P_m$=0.01 | 25.10 | 33 | 40,800 | 436 |
| | $P_m$=0.03 | 28.09 | 51 | 20,800 | 499 |
| | $P_m$=0.05 | 59.22 | 161 | 64,800 | 499 |
| | proposed | 27.66 | 37 | 30,400 | 484 |
| De Jong's f1 (Sphere) | $P_m$=0.00 | 16.98 | 21 | 123,200 | 162 |
| | $P_m$=0.01 | 19.16 | 24 | 100,000 | 216 |
| | $P_m$=0.03 | 26.65 | 38 | 234,000 | 136 |
| | $P_m$=0.05 | 52.06 | 69 | 756,000 | 89 |
| | proposed | 32.50 | 46 | 75,200 | 383 |
| De Jong's f3 (Step) | $P_m$=0.00 | 10.68 | 13 | 11,200 | 493 |
| | $P_m$=0.01 | 10.91 | 14 | 6,000 | 500 |
| | $P_m$=0.03 | 12.94 | 17 | 7,200 | 500 |
| | $P_m$=0.05 | 18.88 | 30 | 12,400 | 500 |
| | proposed | 11.93 | 17 | 7,200 | 500 |
| Shaffer's f6 | $P_m$=0.00 | 13.69 | 18 | 714,400 | 26 |
| | $P_m$=0.01 | 18.12 | 23 | 470,400 | 51 |
| | $P_m$=0.03 | 21.59 | 26 | 896,400 | 34 |
| | $P_m$=0.05 | 62.21 | 32 | 1,491,600 | 39 |
| | proposed | 21.95 | 26 | 313,200 | 87 |
| Rastrigin | $P_m$=0.00 | 15.36 | 20 | 109,200 | 159 |
| | $P_m$=0.01 | 18.55 | 23 | 105,600 | 197 |
| | $P_m$=0.03 | 28.69 | 36 | 222,000 | 157 |
| | $P_m$=0.05 | 75.01 | 92 | 1,078,800 | 96 |
| | proposed | 26.31 | 33 | 81,600 | 318 |
| Schwefel | $P_m$=0.00 | 13.76 | 17 | 36,000 | 333 |
| | $P_m$=0.01 | 15.83 | 27 | 11,200 | 499 |
| | $P_m$=0.03 | 21.16 | 32 | 13,200 | 500 |
| | $P_m$=0.05 | 49.12 | 121 | 48,800 | 500 |
| | proposed | 18.30 | 30 | 12,400 | 496 |
| Griewangk | $P_m$=0.00 | 15.81 | 19 | 624,000 | 32 |
| | $P_m$=0.01 | 26.88 | 30 | 421,600 | 74 |
| | $P_m$=0.03 | 45.65 | 52 | 636,000 | 96 |
| | $P_m$=0.05 | 94.59 | 111 | 2,508,800 | 58 |
| | proposed | 70.99 | 69 | 728,000 | 134 |

Figure 2 and Figure 3 show the adaptive behavior of the proposed diversity control procedure. They are the plots of the number that each preference type is selected to participate in the crossover. The plots of two problems, one-max and multimodal function, are shown. They are the representative of low and high diversity requirement problems respectively. For clarity of the presentation, the data are plotted to the generation 50. They are averaged from 500 runs.

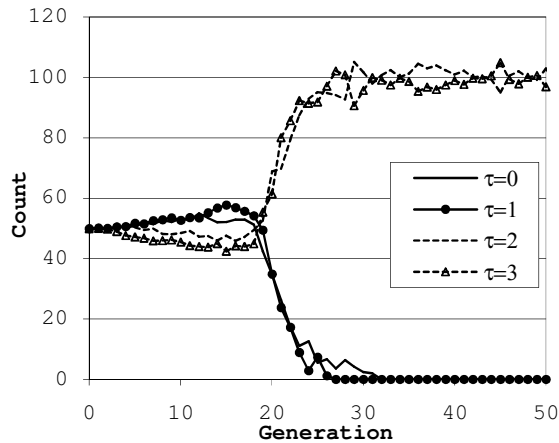

Figure 2. The adaptation of preference type of the one-max problem.

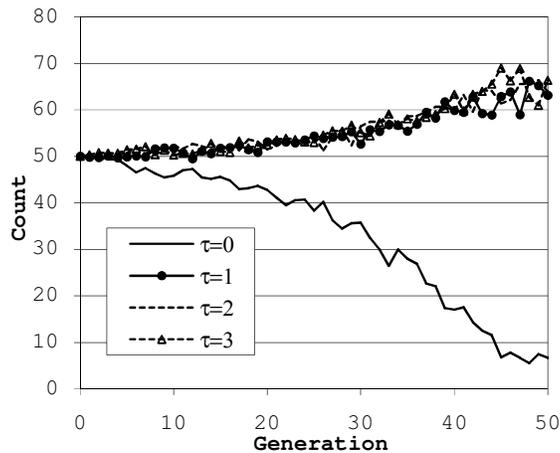

Figure 3. The adaptation of preference type of the multimodal function.

For the one-max problem, the plots show the adaptation toward low diversity. The average generation used of this problem is 13.01 generations. Within this time, the plots show that the lower preference types which are the preference type 0 and 1 are used more frequently. However, once the solution is obtained, the preference type 0 and 1 (prefer less diversity) cause no more distinct solutions (please note that the mutation operator is not used in the experiments therefore there is no other mechanism to generate new genetic materials). Hence they cannot generate any contribution, their use are declined.

For the multimodal function, which is the harder problem and requires high diversity to solve it efficiently, the plots clearly show the adaptation toward more diversity. The preference type 0 does drop rapidly since the early generation.

The diversity plots of traditional GA without mutation versus the proposed diversity control procedure are shown in Figure 4 and Figure 5. The diversity is calculated from the average Hamming distance of all possible pairs of individual in each generation. It is normalized by the length of chromosome as shown in equation 6.

$$Diversity = \frac{\sum_{i=1}^{n} \sum_{j=1}^{n} h(I_i, I_j)}{n^2 \cdot l} \tag{6}$$

$I_i$ and $I_j$ are the $i$th and $j$th individual in the population respectively, $h$ is the Hamming distance of two individuals, $n$ is the population size, and $l$ is the length of chromosome.

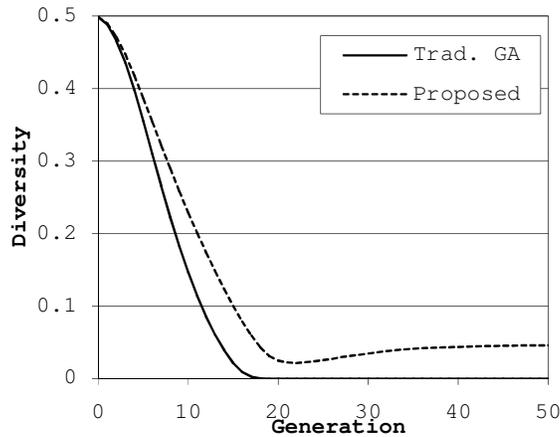

Figure 4. The diversity comparison between the traditional GA and the proposed diversity control procedure of the one-max problem.

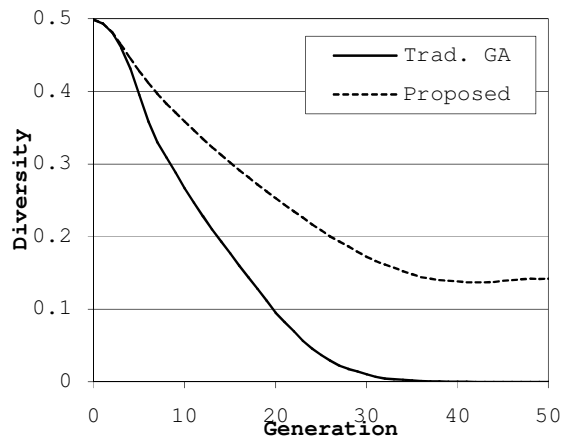

Figure 5. The diversity comparison between the traditional GA and the proposed diversity control procedure of the multimodal function.

The maximum value of the diversity is 0.5 that means the number of bit 0 and 1 in each chromosome position are equal. It usually happens when the population of individual is randomly generated at generation 0. The minimum value of the diversity is 0 that means all individuals in the population are the same.

For one-max problem, the diversity plot of the traditional GA and the proposed diversity control procedure are quite similar since the problem requires only low diversity. Conversely, for multimodal function which is a high diversity requirement problem, the diversity plot shows a high diversity for the proposed procedure. These plots show that the diversity is adapted toward the suitable values for the two problems.

The results demonstrate clearly the ability to adapt the diversity in the population. The proposed method is able to adapt the diversity for a given problem using the preference type and results in the efficient use of resource as can be seen from the computational effort.

Table 4. Performance comparison of the proposed method and the DCGA.

| Problem | Method | AVFE | N | +/- |
|---|---|---|---|---|
| One-Max | $P_m$=0.00, one-point | 3,320.05 | 500 | - |
| | $P_m$=0.03, one-point | 4,935.15 | 500 | - |
| | $P_m$=0.00, two-point | 2,525.95 | 500 | - |
| | proposed | 1,848.33 | 487 | |
| Deceptive function | $P_m$=0.00, one-point | 2,890.51 | 383 | + |
| | $P_m$=0.03, one-point | 10,939.33 | 404 | - |
| | $P_m$=0.00, two-point | 3,182.00 | 403 | + |
| | proposed | 2,005.82 | 163 | |
| Multimodal function | $P_m$=0.00, one-point | 47,038.34 | 47 | - |
| | $P_m$=0.03, one-point | 15,278.04 | 485 | + |
| | $P_m$=0.00, two-point | 19,180.14 | 107 | - |
| | proposed | 3,789.20 | 95 | |
| De Jong's f1 (Sphere) | $P_m$=0.00, one-point | 6,059.00 | 130 | + |
| | $P_m$=0.03, one-point | 7,199.42 | 223 | + |
| | $P_m$=0.00, two-point | 7,419.25 | 122 | + |
| | proposed | 3,044.39 | 49 | |
| De Jong's f3 (Step) | $P_m$=0.00, one-point | 3,386.85 | 416 | - |
| | $P_m$=0.03, one-point | 3,082.60 | 500 | + |
| | $P_m$=0.00, two-point | 2,395.00 | 483 | + |
| | proposed | 2,349.12 | 356 | |
| Shaffer's f6 | $P_m$=0.00, one-point | 2,859.32 | 19 | + |
| | $P_m$=0.03, one-point | 4,617.90 | 63 | + |
| | $P_m$=0.00, two-point | 3,900.54 | 35 | + |
| | proposed | 2,154.71 | 7 | |
| Rastrigin | $P_m$=0.00, one-point | 5,959.18 | 88 | - |
| | $P_m$=0.03, one-point | 7,331.68 | 202 | + |
| | $P_m$=0.00, two-point | 6,585.18 | 115 | + |
| | proposed | 2,367.25 | 36 | |
| Schwefel | $P_m$=0.00, one-point | 13,333.61 | 219 | - |
| | $P_m$=0.03, one-point | 8,050.00 | 500 | - |
| | $P_m$=0.00, two-point | 9,761.00 | 323 | - |
| | proposed | 2,405.54 | 160 | |
| Griewangk | $P_m$=0.00, one-point | 3,794.62 | 13 | + |
| | $P_m$=0.03, one-point | 28,336.78 | 90 | + |
| | $P_m$=0.00, two-point | 6,052.19 | 21 | + |
| | proposed | 5,499.70 | 10 | |

# 7. COMPARE TO DCGA

Since DCGA is one of the most well-known methods to maintain population diversity in GA, it is compared with the proposed method. Because the number of individuals processed in each generation of DCGA is not constant, another performance measurement is used. The metric is calculated from the quotient of average function evaluation and probability of success (from 500 runs). The higher value is the lower performance.

Table 4 shows the performance comparison between DCGA and the proposed method. The average function evaluation (denoted by AVFE) is the average number of fitness evaluation used of the success runs (from 500 runs). The "+/-" column determine the performance of DCGA in any configuration comparing with the proposed method. The "+" sign determine the superior performance of DCGA, and vice versa, the "-" sign determine the inferior performance.

The DCGA is set to 3 configurations which are $P_m$=0.00 and one-point crossover used, $P_m$=0.03 and one-point crossover used, and $P_m$=0.00 and two-point crossover used. The other parameters used in this experiment are the same as shown in Table 2. The objective is to investigate the change of performance of DCGA when the difference configurations are set.

The results from Table 4 show that DCGA is superior to the proposed method for some problems and inferior for some problems too. Summarily, the performance of the two methods is comparable. However, the different configurations of DCGA lead to large variations in performance. This means DCGA is very sensitive to parameter setting. The wrong configuration leads to the poor performance. The proposed method is superior to DCGA in this viewpoint since it can adapt itself to fit the problem automatically without tuning the parameters.

# 8. CONCLUSION

In solving problems using GA, the diversity maintenance of the population is an important issue. Most method that incorporates the diversity maintenance in GA requires the knowledge of the suitable degree of diversity in the problems to set parameters for the run correctly. This work proposed an adaptive procedure to automatically adjust a suitable degree of diversity in the population for a given problem. The procedure works concurrently with the GA search. The main mechanism of the proposed procedure is the competition of different groups with different *preference type*. Multiple preference types are used in the selection process of GA and the adaptive procedure is employed to make best use of a specific preference type that is suitable for the problem at hand. The self-adaptation procedure uses *contribution* as a measurement of the success of each preference type in solving the problem effectively and adapts towards more use of that preference type. This gives rise to the adaptive behavior of the proposed procedure.

The proposed procedure is tested with the well-known test problems in GA. The problems are varied in the requirement of the diversity of the population to solve the problems effectively. From the experiment, the adaptive procedure works successfully for the standard test problems of GA. It has capability to adapt the suitable diversity for the different problems. Moreover, the performance of the adaptive method is comparable to both the non-adaptive method that has the correct parameter setting for the given problem and the DCGA.

From this experiment, the proposed method shows the advantage in the term of flexibility of use. The proposed method can adapt the diversity of the population for a given problem without the knowledge of correct parameter setting and it has a good performance in finding the solution.

## ACKNOWLEDGEMENT

The first author would like to acknowledge the support of the Royal Golden Jubilee Ph.D. Graduates Program by Thailand Research Fund organization.

**Authors**

**Chaiwat Jassadapakorn** earned B.Sc. in Computer Science from Silpakorn University, Thailand in 1995 and M.Sc. in Computer Science from Chulalongkorn University, Thailand in 1998. Presently, he is a Ph.D. candidate in the department of computer engineering, Chulalongkorn University.

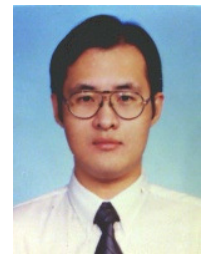

**Prabhas Chongstitvatana** earned B.Eng. in Electrical Engineering from Kasetsart University, Thailand in 1980 and Ph.D. from the department of artificial intelligence, Edinburgh University, U.K. in 1992. Presently, he is a professor in the department of computer engineering, Chulalongkorn University. His research included robotics, evolutionary computation and computer architecture. The current work involves bioinformatics and grid computing. He is actively promote the collaboration to create Thai national grid for scientific computing. He is the member of Thailand Engineering Institute, Thai Academy of Science and Technology, Thai Robotics Society, Thai Embedded System Association, ECTI Association of Thailand and IEEE Robotics and Automation Society.

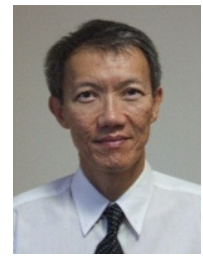

# APPENDIX

This section shows the graphical plots of the seven problems used in the experiment.

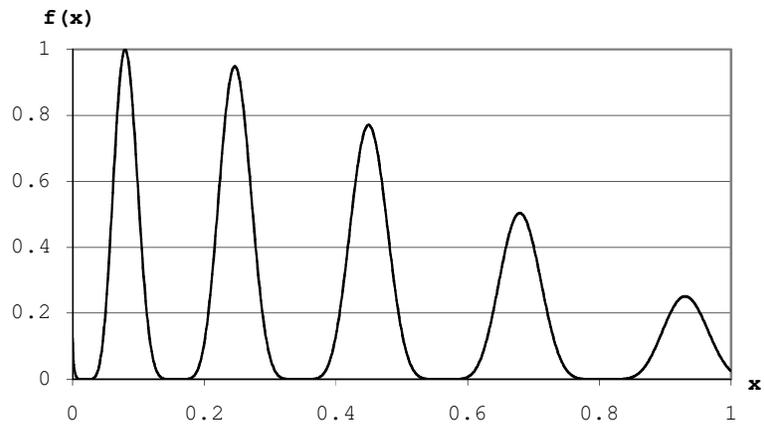

Figure 6. The plot of multimodal function.

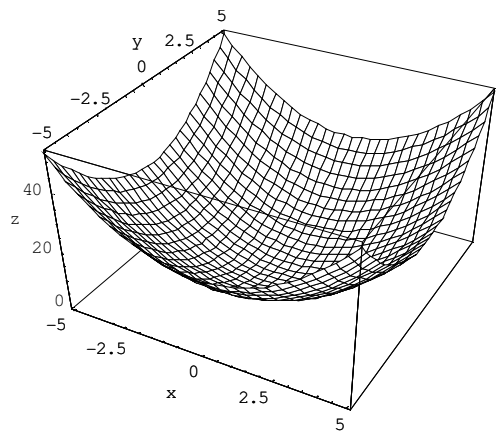

Figure 7. The plot of 2-D De Jong's f1 (Sphere) function.

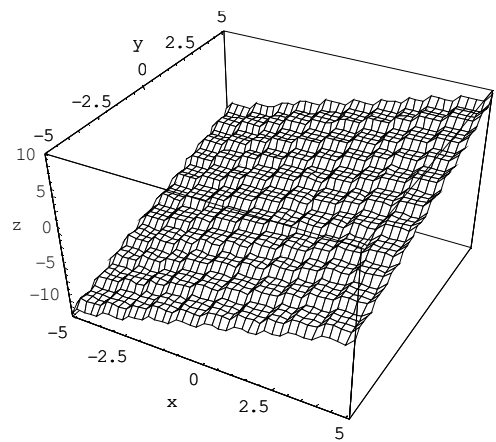

Figure 8. The plot of 2-D De Jong's f3 (Step) function.

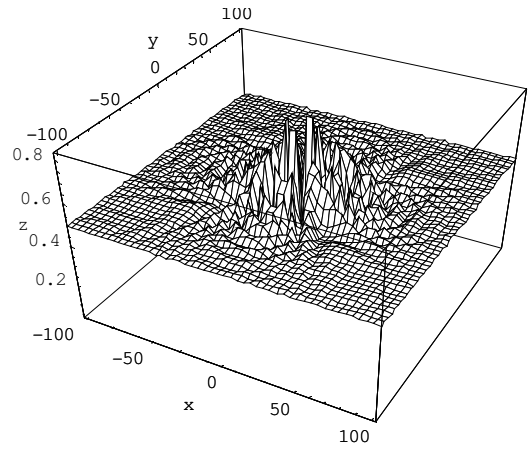

Figure 9. The plot of 2-D Shaffer's f6 function.

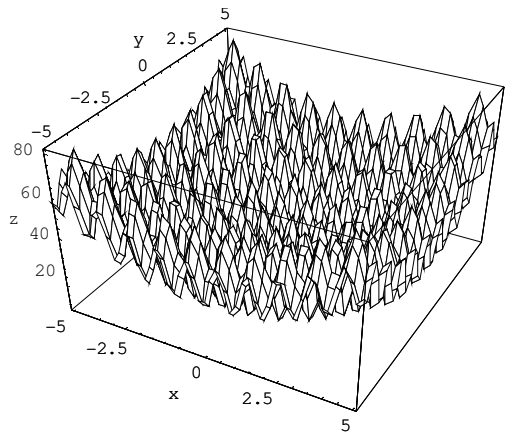

Figure 10. The plot of 2-D Rastrigin function.

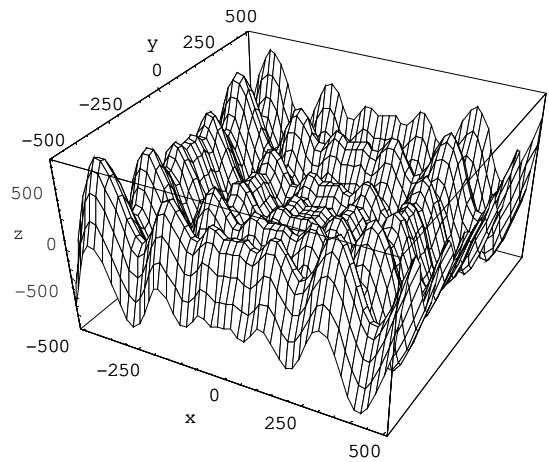

Figure 11. The plot of 2-D Schwefel function.

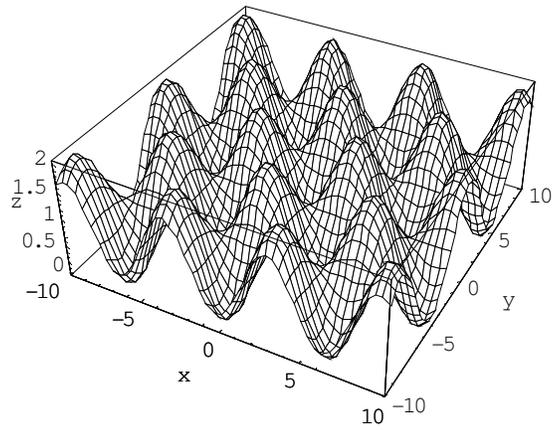

Figure 12. The plot of 2-D (zoom) Griewangk function